\documentclass{article} % For LaTeX2e
\usepackage{iclr2015,times}
\usepackage{hyperref}
\usepackage{url}

\usepackage{amsmath,bm}
\usepackage{amssymb}
\usepackage{algorithm}
\usepackage{algorithmic}
\usepackage{graphicx}
\usepackage{tabularx}
\usepackage{array,arydshln}
\usepackage{subcaption}

\makeatletter
\newcommand*{\rom}[1]{\expandafter\@slowromancap\romannumeral #1@}
\makeatother

\title{Deep Captioning with Multimodal Recurrent Neural Networks (m-RNN)}

\author{
Junhua Mao\\
University of California, Los Angeles; Baidu Research \\
\texttt{mjhustc@ucla.edu} \\
\And
Wei Xu \& Yi Yang \& Jiang Wang \& Zhiheng Huang\\
Baidu Research \\
\texttt{\{wei.xu,yangyi05,wangjiang03,huangzhiheng\}@baidu.com} \\
\AND
Alan Yuille \\
University of California, Los Angeles \\
\texttt{yuille@stat.ucla.edu} 
}

\iclrfinalcopy % Uncomment for camera-ready version

\iclrconference % Uncomment if submitted as conference paper instead of workshop

\begin{document}

\maketitle

\begin{abstract}

In this paper, we present a multimodal Recurrent Neural Network (m-RNN) model for generating novel image captions.
It directly models the probability distribution of generating a word given previous words and an image.
Image captions are generated according to this distribution.
The model consists of two sub-networks: a deep recurrent neural network for sentences and a deep convolutional network for images. 
These two sub-networks interact with each other in a multimodal layer to form the whole m-RNN model.
The effectiveness of our model is validated on four benchmark datasets (IAPR TC-12, Flickr 8K, Flickr 30K and MS COCO).
Our model outperforms the state-of-the-art methods.
In addition, we apply the m-RNN model to retrieval tasks for retrieving images or sentences, and achieves significant performance improvement over the state-of-the-art methods which directly optimize the ranking objective function for retrieval.
The project page of this work is: \url{www.stat.ucla.edu/~junhua.mao/m-RNN.html}.
\footnote{Most recently, we adopt a simple strategy to boost the performance of image captioning task significantly.
More details are shown in Section \ref{sec:NNref}.
The code and related data (e.g. refined image features and hypotheses sentences generated by the m-RNN model) are available at \url{https://github.com/mjhucla/mRNN-CR}.}

\end{abstract}

\section{Introduction}

Obtaining sentence level descriptions for images is becoming an important task and it has many applications, such as early childhood education, image retrieval, and navigation for the blind.
Thanks to the rapid development of computer vision and natural language processing technologies, recent work has made significant progress on this task (see a brief review in Section \ref{sec:related_work}).
Many previous methods treat it as a retrieval task.
They learn a joint embedding to map the features of both sentences and images to the same semantic space.
These methods generate image captions by retrieving them from a sentence database.
Thus, they lack the ability of generating novel sentences or describing images that contain novel combinations of objects and scenes.

In this work, we propose a multimodal Recurrent Neural Networks (m-RNN) model \footnote{A previous version of this work appears in the NIPS 2014 Deep Learning Workshop with the title ``Explain Images with Multimodal Recurrent Neural Networks'' \url{http://arxiv.org/abs/1410.1090} (\cite{mao2014explain}). We observed subsequent arXiv papers which also use recurrent neural networks in this topic and cite our work. We gratefully acknowledge them.} to address both the task of generating novel sentences descriptions for images, and the task of image and sentence retrieval.
The whole m-RNN model contains a language model part, a vision part and a multimodal part.
The language model part learns a dense feature embedding for each word in the dictionary and stores the semantic temporal context in recurrent layers.
The vision part contains a deep Convolutional Neural Network (CNN) which generates the image representation.
The multimodal part connects the language model and the deep CNN together by a one-layer representation.
Our m-RNN model is learned using a log-likelihood cost function (see details in Section \ref{sec:trainCost}).
The errors can be backpropagated to the three parts of the m-RNN model to update the model parameters simultaneously.
% To the best of our knowledge, this is the first work that incorporates the Recurrent Neural Network in a deep multimodal architecture.

In the experiments, we validate our model on four benchmark datasets: IAPR TC-12 (\cite{grubinger2006iapr}), Flickr 8K (\cite{rashtchian2010collecting}), Flickr 30K (\cite{hodoshimage}) and MS COCO (\cite{lin2014microsoft}).
We show that our method achieves state-of-the-art performance, significantly outperforming all the other methods for the three tasks: generating novel sentences, retrieving images given a sentence and retrieving sentences given an image.
Our framework is general and can be further improved by incorporating more powerful deep representations for images and sentences.

\begin{figure}[tb!]
\begin{center}
\includegraphics[width=0.98\linewidth]{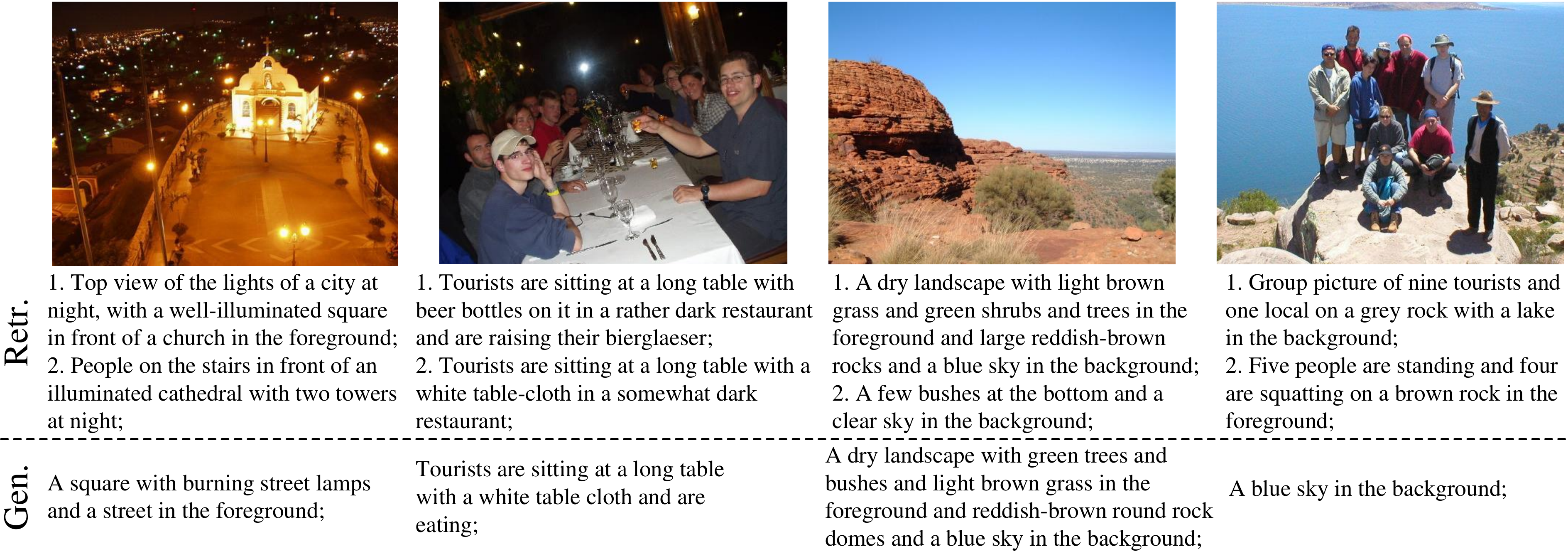}
\end{center}
\vspace{-0.3cm}
   \caption{Examples of the generated and two top-ranked retrieved sentences given the query image from IAPR TC-12 dataset.
   The sentences can well describe the content of the images.
   We show a failure case in the fourth image, where the model mistakenly treats the lake as the sky and misses all the people.
   More examples from the MS COCO dataset can be found on the project page: \url{www.stat.ucla.edu/~junhua.mao/m-RNN.html}.
   }
\vspace{-0.3cm}
\label{fig:res_example}
\end{figure}

\section{Related Work}
\label{sec:related_work}

\textbf{Deep model for computer vision and natural language.}
The methods based on the deep neural network developed rapidly in recent years in both the field of computer vision and natural language.
For computer vision, \cite{krizhevsky2012imagenet} propose a deep Convolutional Neural Networks (CNN) with 8 layers (denoted as AlexNet) and outperform previous methods by a large margin in the image classification task of ImageNet challenge (\cite{ILSVRCarxiv14}).
This network structure is widely used in computer vision, e.g. \cite{girshick2014rcnn} design a object detection framework (RCNN) based on this work.
Recently, \cite{simonyan2014very} propose a CNN with over 16 layers (denoted as VggNet) and performs substantially better than the AlexNet.
For natural language, the Recurrent Neural Network (RNN) shows the state-of-the-art performance in many tasks, such as speech recognition and word embedding learning (\cite{mikolov2010recurrent,mikolov2011extensions,mikolov2013distributed}).
Recently, RNNs have been successfully applied to machine translation to extract semantic information from the source sentence and generate target sentences (e.g. \cite{kalchbrenner2013recurrent}, \cite{cho2014learning} and \cite{sutskever2014sequence}).

\textbf{Image-sentence retrieval.}
Many previous methods treat the task of describing images as a retrieval task and formulate the problem as a ranking or embedding learning problem (\cite{hodosh2013framing,frome2013devise,socher2014grounded}).
They first extract the word and sentence features (e.g. \cite{socher2014grounded} uses dependency tree Recursive Neural Network to extract sentence features) as well as the image features.
Then they optimize a ranking cost to learn an embedding model that maps both the sentence feature and the image feature to a common semantic feature space.
In this way, they can directly calculate the distance between images and sentences.
Recently, \cite{karpathy2014fragment} show that object level image features based on object detection results can generate better results than image features extracted at the global level.

\textbf{Generating novel sentence descriptions for images.}
There are generally three categories of methods for this task.
The first category assumes a specific rule of the language grammar.
They parse the sentence and divide it into several parts (\cite{mitchell2012midge,gupta2012image}).
Each part is associated with an object or an attribute in the image (e.g. \cite{kulkarni2011baby} uses a Conditional Random Field model and \cite{farhadi2010every} uses a Markov Random Field model).
This kind of method generates sentences that are syntactically correct.
The second category retrieves similar captioned images, and generates new descriptions by generalizing and re-composing the retrieved captions (\cite{kuznetsova2014treetalk}).
The third category of methods, which is more related to our method, learns a probability density over the space of multimodal inputs (i.e. sentences and images), using for example, Deep Boltzmann Machines (\cite{srivastava2012multimodal}), and topic models (\cite{barnard2003matching,jia2011learning}).
They generate sentences with richer and more flexible structure than the first group.
The probability of generating sentences using the model can serve as the affinity metric for retrieval.
Our method falls into this category.
More closely related to our tasks and method is the work of \cite{kiros2013multimodal}, which is built on a Log-BiLinear model (\cite{mnih2007three}) and use AlexNet to extract visual features.
It needs a fixed length of context (i.e. five words), whereas in our model, the temporal context is stored in a recurrent architecture, which allows arbitrary context length.

Shortly after \cite{mao2014explain}, several papers appear with record breaking results   (e.g. \cite{kiros2014unifying,karpathy2014deep,vinyals2014show,donahue2014long,fang2014captions,chen2014learning}).
Many of them are built on recurrent neural networks.
It demonstrates the effectiveness of storing context information in a recurrent layer.
Our work has two major difference from these methods.
Firstly, we incorporate a two-layer word embedding system in the m-RNN network structure which learns the word representation more efficiently than the single-layer word embedding.
Secondly, we do not use the recurrent layer to store the visual information.
The image representation is inputted to the m-RNN model along with every word in the sentence description.
It utilizes of the capacity of the recurrent layer more efficiently, and allows us to achieve state-of-the-art performance using a relatively small dimensional recurrent layer.
In the experiments, we show that these two strategies lead to better performance.
Our method is still the best-performing approach for almost all the evaluation metrics.
% Thirdly, the activation of the word embedding layer directly affect the multimodal layer.
% In the experiments, we show that all of these strategies lead to a better performance.

\vspace{-0.2cm}
\section{Model Architecture}
\vspace{-0.2cm}

\begin{figure}[tb!]
\begin{center}
\includegraphics[width=0.9\linewidth]{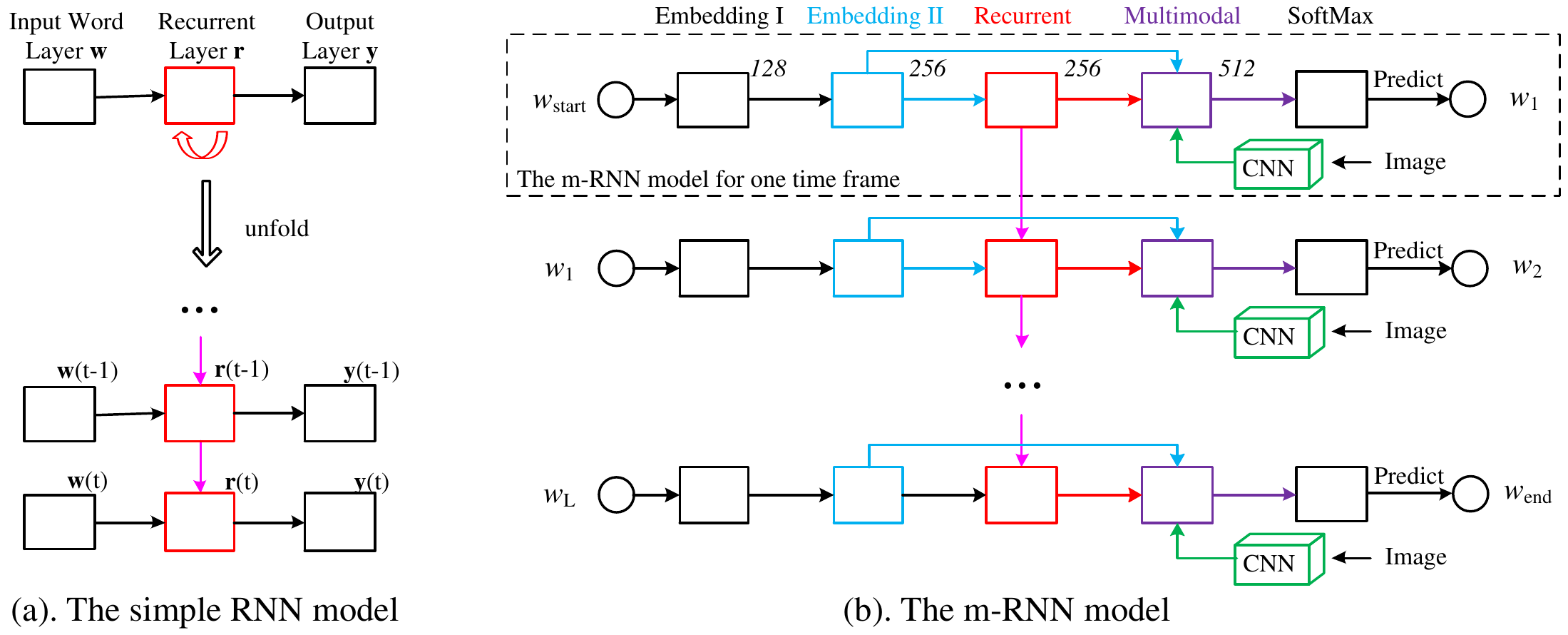}
\end{center}
\vspace{-0.3cm}
   \caption{Illustration of the simple Recurrent Neural Network (RNN) and our multimodal Recurrent Neural Network (m-RNN) architecture.
   (a). The simple RNN. 
   (b). Our m-RNN model.
   The inputs of our model are an image and its corresponding sentence descriptions.
   $w_\text{1}$, $w_\text{2}$, ..., $w_\text{L}$ represents the words in a sentence.
   We add a start sign $w_\text{start}$ and an end sign $w_\text{end}$ to all the training sentences.
   The model estimates the probability distribution of the next word given previous words and the image.
   It consists of five layers (i.e. two word embedding layers, a recurrent layer, a multimodal layer and a softmax layer) and a deep CNN in each time frame.
   The number above each layer indicates the dimension of the layer.
   The weights are shared among all the time frames.
   (Best viewed in color)
   % This architecture is much deeper than the simple RNN.
   % of structure without considering the extendibility of the recurrent layer.
   % (c). The illustration the unfolded m-RNN. 
   % We can unfold the recurrent layer, which leads to the temporal depth of the network.
   % The model parameters are shared for each temporal frame of the m-RNN model.
   }
\vspace{-0.3cm}
\label{fig:illu_RNN}
\end{figure}

% For layer \textcircled{\raisebox{-0.9pt}{1}} and layer \textcircled{2},
\subsection{Simple recurrent neural network}
\label{sec:sRNN}
We briefly introduce the simple Recurrent Neural Network (RNN) or Elman network (\cite{elman1990finding}).
Its architecture is shown in Figure \ref{fig:illu_RNN}(a).
It has three types of layers in each time frame: the input word layer $\mathbf{w}$, the recurrent layer $\mathbf{r}$ and the output layer $\mathbf{y}$.
The activation of input, recurrent and output layers at time $t$ is denoted as $\mathbf{w}(t)$, $\mathbf{r}(t)$, and $\mathbf{y}(t)$ respectively.
$\mathbf{w}(t)$ denotes the current word vector, which can be a simple 1-of-N coding representation $\mathbf{h}(t)$ (i.e. the one-hot representation, which is binary and has the same dimension as the vocabulary size with only one non-zero element) \cite{mikolov2010recurrent}.
$\mathbf{y}(t)$ can be calculated as follows:
\begin{equation}
\mathbf{x}(t) = [\mathbf{w}(t)\ \ \mathbf{r}(t-1)];\ \ \ 
\mathbf{r}(t)=f_1(\mathbf{U} \cdot \mathbf{x}(t));\ \ \ 
\mathbf{y}(t)=g_1(\mathbf{V} \cdot \mathbf{r}(t));
\end{equation}
where $\mathbf{x}(t)$ is a vector that concatenates $\mathbf{w}(t)$ and $\mathbf{r}(t-1)$, $f_1(.)$ and $g_1(.)$ are element-wise sigmoid and softmax function respectively, and $\mathbf{U}$, $\mathbf{V}$ are weights which will be learned.

The size of the RNN is adaptive to the length of the input sequence.
The recurrent layers connect the sub-networks in different time frames.
Accordingly, when we do backpropagation, we need to propagate the error through recurrent connections back in time (\cite{rumelhart1988learning}).

\subsection{Our m-RNN model}
The structure of our multimodal Recurrent Neural Network (m-RNN) is shown in Figure \ref{fig:illu_RNN}(b).
% The m-RNN model is much deeper than the simple RNN model.
It has five layers in each time frame: two word embedding layers, the recurrent layer, the multimodal layer, and the softmax layer).

The two word embedding layers embed the one-hot input into a dense word representation.
It encodes both the syntactic and semantic  meaning of the words.
% It has several advantages.
% Firstly, it significantly lowers the number of parameters in the networks because the dense word vector (128 dimension) is much smaller than the one-hot word vector.
% Secondly, the dense word embedding encodes the semantic meaning of the words (\cite{mikolov2013efficient}).
The semantically relevant words can be found by calculating the Euclidean distance between two dense word vectors in embedding layers.
Most of the sentence-image multimodal models (\cite{karpathy2014fragment,frome2013devise,socher2014grounded,kiros2013multimodal}) use pre-computed word embedding vectors as the initialization of their model. 
In contrast, we randomly initialize our word embedding layers and learn them from the training data.
We show that this random initialization is sufficient for our architecture to generate the state-of-the-art result.
% To further refine the word representation, we add a hidden layer after the initial word embedding layer and 
We treat the activation of the word embedding layer \rom{2} (see Figure \ref{fig:illu_RNN}(b)) as the final word representation, which is one of the three direct inputs of the multimodal layer.

After the two word embedding layers, we have a recurrent layer with 256 dimensions.
The calculation of the recurrent layer is slightly different from the calculation for the simple RNN.
Instead of concatenating the word representation at time $t$ (denoted as $\mathbf{w}(t)$) and the recurrent layer activation at time $t-1$ (denoted as $\mathbf{r}(t-1)$), we first map $\mathbf{r}(t-1)$ into the same vector space as $\mathbf{w}(t)$ and add them together:
\begin{equation}
\mathbf{r}(t)=f_2(\mathbf{U}_r \cdot \mathbf{r}(t-1) + \mathbf{w}(t));
\end{equation}
% where $\mathbf{s}$ and $\mathbf{w}$ denotes the recurrent layer vector and the word representation respectively.
% This strategy will reduce the number of parameters and accelerate the training and testing process.
where ``+'' represents element-wise addition.
We set $f_2(.)$ to be the Rectified Linear Unit (ReLU), inspired by its the recent success when training very deep structure in computer vision field (\cite{nair2010rectified,krizhevsky2012imagenet}).
This differs from the simple RNN where the sigmoid function is adopted (see Section \ref{sec:sRNN}).
ReLU is faster, and harder to saturate or overfit the data than non-linear functions like the sigmoid.
When the backpropagation through time (BPTT) is conducted for the RNN with sigmoid function, the vanishing or exploding gradient problem appears since even the simplest RNN model can have a large temporal depth \footnote{We tried Sigmoid and Scaled Hyperbolic Tangent function as the non-linear functions for RNN in the experiments but they lead to the gradient explosion problem easily.}.
Previous work (\cite{mikolov2010recurrent,mikolov2011extensions}) use heuristics, such as the truncated BPTT, to avoid this problem.
The truncated BPTT stops the BPTT after $k$ time steps, where $k$ is a hand-defined hyperparameter.
Because of the good properties of ReLU, we do not need to stop the BPTT at an early stage, which leads to better and more efficient utilization of the data than the truncated BPTT.

After the recurrent layer, we set up a 512 dimensional multimodal layer that connects the language model part and the vision part of the m-RNN model (see Figure \ref{fig:illu_RNN}(b)).
This layer has three inputs: the word-embedding layer \rom{2}, the recurrent layer and the image representation.
For the image representation, here we use the activation of the $\text{7}^\text{th}$ layer of AlexNet (\cite{krizhevsky2012imagenet}) or $\text{15}^\text{th}$ layer of VggNet (\cite{simonyan2014very}), though our framework can use any image features. 
% The language model part includes the word embedding layer 2 (the final word representation) and the recurrent layer (the sentence context).
% The vision part contains the image feature extraction network.
% Here we connect the seventh layer of AlexNet \cite{krizhevsky2012imagenet} or VGGNet \cite{simonyan2014very} to the multimodal layer (please refer to Section \ref{sec:ImgSenFeat} for more details), although our framework can use any image features.
We map the activation of the three layers to the same multimodal feature space and add them together to obtain the activation of the multimodal layer:
\begin{equation}
\mathbf{m}(t)=g_2(\mathbf{V}_w \cdot \mathbf{w}(t) + \mathbf{V}_r \cdot \mathbf{r}(t) + \mathbf{V}_I \cdot \mathbf{I});
\label{equ:multimodal}
\end{equation}
where ``+'' denotes element-wise addition, $\mathbf{m}$ denotes the multimodal layer feature vector, $\mathbf{I}$ denotes the image feature.
$g_2(.)$ is the element-wise scaled hyperbolic tangent function (\cite{lecun2012efficient}):
\begin{equation}
g_2(x) = 1.7159 \cdot \tanh( \frac{2}{3} x)
\end{equation}
This function forces the gradients into the most non-linear value range and leads to a faster training process than the basic hyperbolic tangent function.

%We do not restrict the norm the three feature layers that connect to the multimodal layer.
%Our experiments shows that the L2 norm of the hidden layer, 

Both the simple RNN and m-RNN models have a softmax layer that generates the probability distribution of the next word.
The dimension of this layer is the vocabulary size $M$, which is different for different datasets.

\section{Training the m-RNN}
\label{sec:trainCost}
To train our m-RNN model we adopt a log-likelihood cost function.
It is related to the \emph{Perplexity} of the sentences in the training set given their corresponding images.
Perplexity is a standard measure for evaluating language model.
The perplexity for one word sequence (i.e. a sentence) $w_{1:L}$ is calculated as follows:
\begin{equation}
\log_2 \mathcal{PPL}(w_{1:L}|\mathbf{I}) = -\frac{1}{L} \sum_{n=1}^{L} \log_2 P(w_n|w_{1:n-1},\mathbf{I})
\label{equ:perp}
\end{equation}
where $L$ is the length of the word sequence, $\mathcal{PPL}(w_{1:L}|\mathbf{I})$ denotes the perplexity of the sentence $w_{1:L}$ given the image $\mathbf{I}$.
$P(w_n|w_{1:n-1},\mathbf{I})$ is the probability of generating the word $w_n$ given $\mathbf{I}$ and previous words $w_{1:n-1}$.
It corresponds to the activation of the SoftMax layer of our model.

The cost function of our model is the average log-likelihood of the words given their context words and corresponding images in the training sentences plus a regularization term.
It can be calculated by the perplexity:
\begin{equation}
\mathcal{C} = \frac{1}{N} \sum_{i=1}^{N_s} L_i \cdot \log_2 \mathcal{PPL}(w_{1:L_i}^{(i)}|\mathbf{I}^{(i)}) + \lambda_\theta \cdot \left \| \theta \right \|_2^2
\end{equation}
where $N_s$ and $N$ denotes the number of sentences and the number of words in the training set receptively, $L_i$ denotes the length of $i^{th}$ sentences, and $\theta$ represents the model parameters.
% It is equivalent to the reciprocal of the geometric mean of the probability to generate the training sentences using the model.

Our training objective is to minimize this cost function, which is equivalent to maximize the probability of generating the sentences in the training set using the model. % given their corresponding images.
The cost function is differentiable and we use backpropagation to learn the model parameters.

\section{Sentence Generation, Image Retrieval and Sentence Retrieval}
We use the trained m-RNN model for three tasks: 1) Sentences generation, 2) Image retrieval (retrieving most relevant images to the given sentence), 3) Sentence retrieval (retrieving most relevant sentences to the given image).

The sentence generation process is straightforward.
Starting from the start sign $w_\text{start}$ or arbitrary number of reference words (e.g. we can input the first K words in the reference sentence to the model and then start to generate new words), our model can calculate the probability distribution of the next word: $P(w_n|w_{1:n-1},\mathbf{I})$.
Then we can sample from this probability distribution to pick the next word.
In practice, we find that selecting the word with the maximum probability performs slightly better than sampling.
After that, we input the picked word to the model and continue the process until the model outputs the end sign $w_\text{end}$.

For the retrieval tasks, we use our model to calculate the probability of generating a sentence $w_{1:L}$ given an image $\mathbf{I}$: $P(w_{1:L}|\mathbf{I})=\prod_n P(w_n|w_{1:n-1},\mathbf{I})$.
The probability can be treated as an affinity measurement between sentences and images.

For the image retrieval task, given the query sentence $w_{1:L}^{Q}$, we rank the dataset images $\mathbf{I}^{D}$ according to the probability $P(w_{1:L}^{Q}|\mathbf{I}^{D})$ and retrieved the top ranked images.
This is equivalent to the perplexity-based image retrieval in \cite{kiros2013multimodal}. 

The sentence retrieval task is trickier because there might be some sentences that have high probability or perplexity for any image query (e.g. sentences consist of many frequently appeared words).
To solve this problem, \cite{kiros2013multimodal} uses the perplexity of a sentence conditioned on the averaged image feature across the training set as the reference perplexity to normalize the original perplexity.
Different from them, we use the \emph{normalized probability} where the normalization factor is the marginal probability of $w_{1:L}^{D}$:
\begin{equation}
\vspace{-0.1cm}
P(w_{1:L}^{D}|\mathbf{I}^{Q}) / P(w_{1:L}^{D});
\ \ \ \ \ 
P(w_{1:L}^{D}) = {\textstyle \sum_{\mathbf{I^{'}}}} P(w_{1:L}^{D}|\mathbf{I^{'}}) \cdot P(\mathbf{I^{'}})
\end{equation}
where $w_{1:L}^{D}$ denotes the sentence in the dataset, $\mathbf{I}^{Q}$ denotes the query image, and $\mathbf{I^{'}}$ are images sampled from the training set.
We approximate $P(\mathbf{I^{'}})$ by a constant and ignore this term.
% $P(w_{1:L}|\mathbf{I}) = \mathcal{PPL}(w_{1:L}|\mathbf{I}) ^ {-L}$.
This strategy leads to a much better performance than that in \cite{kiros2013multimodal} in the experiments.
The normalized probability is equivalent to the probability $P(\mathbf{I}^{Q}|w_{1:L}^{D})$, which is symmetric to the probability $P(w_{1:L}^{Q}|\mathbf{I}^{D})$ used in the image retrieval task.

\section{Learning of Sentence and Image Features}
\label{sec:ImgSenFeat}

The architecture of our model allows the gradients from the loss function to be backpropagated to both the language modeling part (i.e. the word embedding layers and the recurrent layer) and the vision part (e.g. the AlexNet or VggNet).

For the language part, as mentioned above, we randomly initialize the language modeling layers and learn their parameters.
For the vision part, we use the pre-trained AlexNet (\cite{krizhevsky2012imagenet}) or the VggNet (\cite{simonyan2014very}) on ImageNet dataset (\cite{ILSVRCarxiv14}).
% The same strategy is used by previous multimodal methods (\cite{kiros2013multimodal,frome2013devise,karpathy2014fragment,socher2014grounded}).
Recently, \cite{karpathy2014fragment} show that using the RCNN object detection results (\cite{girshick2014rcnn}) combined with the AlexNet features performs better than simply treating the image as a whole frame.
In the experiments, we show that our method performs much better than \cite{karpathy2014fragment} when the same image features are used, and is better than or comparable to their results even when they use more sophisticated features based on object detection.

We can update the CNN in the vision part of our model according to the gradient backpropagated from the multimodal layer.
In this paper, we fix the image features and the deep CNN network in the training stage due to a shortage of data.
In future work, we will apply our method on large datasets (e.g. the complete MS COCO dataset, which has not yet been released) and finetune the parameters of the deep CNN network in the training stage.

The m-RNN model is trained using Baidu's internal deep learning platform PADDLE, which allows us to explore many different model architectures in a short period.
The hyperparameters, such as layer dimensions and the choice of the non-linear activation functions, are tuned via cross-validation on Flickr8K dataset and are then fixed across all the experiments.
It takes 25 ms on average to generate a sentence (excluding image feature extraction stage) on a single core CPU.

\section{Experiments}

\subsection{Datasets}
We test our method on four benchmark datasets with sentence level annotations: IAPR TC-12 (\cite{grubinger2006iapr}), Flickr 8K (\cite{rashtchian2010collecting}), Flickr 30K (\cite{hodoshimage}) and MS COCO (\cite{lin2014microsoft}).

% Here are some statistics and our experimental settings for the three datasets:

\textbf{IAPR TC-12}.
This dataset consists of around 20,000 images taken from different locations around the world.
It contains images of different sports and actions, people, animals, cities, landscapes, etc.
For each image, it provides at least one sentence annotation.
On average, there are about 1.7 sentence annotations for one image.
We adopt the standard separation of training and testing set as previous works (\cite{GVS10a,kiros2013multimodal}) with 17,665 images for training and 1962 images for testing.

\textbf{Flickr8K}.
This dataset consists of 8,000 images extracted from Flickr.
For each image, it provides five sentence annotations.
% The grammar of the annotations for this dataset is simpler than that for the IAPR TC-12 dataset.
We adopt the standard separation of training, validation and testing set provided by the dataset.
There are 6,000 images for training, 1,000 images for validation and 1,000 images for testing.

\textbf{Flickr30K}.
This dataset is a recent extension of Flickr8K.
For each image, it also provides five sentences annotations.
It consists of 158,915 crowd-sourced captions describing 31,783 images.
The grammar and style for the annotations of this dataset is similar to Flickr8K.
We follow the previous work (\cite{karpathy2014fragment}) which used 1,000 images for testing.
This dataset, as well as the Flick8K dataset, were originally used for the image-sentence retrieval tasks.
%and there is not public available results of methods for generating novel sentence descriptions.

\textbf{MS COCO}.
The current release of this recently proposed dataset contains 82,783 training images and 40,504 validation images.
For each image, it provides five sentences annotations.
We randomly sampled 4,000 images for validation and 1,000 images for testing from their currently released validation set.
The dataset partition of MS COCO and Flickr30K is available in the project page \footnote{\url{www.stat.ucla.edu/~junhua.mao/m-RNN.html}}.

\subsection{Evaluation metrics}
\label{sec:EvaRet}

\textbf{Sentence Generation}.
Following previous works, we use the sentence perplexity (see Equ. \ref{equ:perp}) and BLEU scores (i.e. B-1, B-2, B-3, and B-4) (\cite{papineni2002bleu}) as the evaluation metrics.
BLEU scores were originally designed for automatic machine translation where they rate the quality of a translated sentences given several reference sentences.
Similarly, we can treat the sentence generation task as the ``translation'' of the content of images to sentences.
BLEU remains the standard evaluation metric for sentence generation methods for images, though it has drawbacks.
For some images, the reference sentences might not contain all the possible descriptions in the image and BLEU might penalize some correctly generated sentences.
Please see more details of the calculation of BLEU scores for this task in the supplementary material section \ref{supp:bleu} 
\footnote{The BLEU outputted by our implementation is slightly lower than the recently released MS COCO caption evaluation toolbox (\cite{capeval2015}) because of different tokenization methods of the sentences. 
We re-evaluate our method using the toolbox in the current version of the paper.}.
% To conduct a fair comparison, we adopt the same sentence generation steps and experiment settings as \cite{kiros2013multimodal}, and generate as many words as there are in the reference sentences to calculate BLEU.
% Note that our model does not need to know the length of the reference sentence because we add an end sign "\#\#END\#\#" at the end of every training sentences and we can stop the generation process when our model outputs the end sign.

\textbf{Sentence Retrieval and Image Retrieval}. 
% For Flickr8K and Flickr30K datasets, 
We adopt the same evaluation metrics as previous works (\cite{socher2014grounded,frome2013devise,karpathy2014fragment}) for both the tasks of sentences retrieval and image retrieval.
We use R@K (K = 1, 5, 10) as the measurement.
R@K is the recall rate of a correctly retrieved groundtruth given top K candidates.
% the first retrieved groundtruth sentences (sentence retrieval task) or images (image retrieval task).
Higher R@K usually means better retrieval performance.
Since we care most about the top-ranked retrieved results, the R@K scores with smaller K are more important.

The Med r is another metric we use, which is the median rank of the first retrieved groundtruth sentence or image.
Lower Med r usually means better performance.
For IAPR TC-12 datasets, we use additional evaluation metrics to conduct a fair comparison with previous work (\cite{kiros2013multimodal}).
Please see the details in the supplementary material section \ref{supp:bleu}.

\subsection{Results on IAPR TC-12}
The results of the sentence generation task\footnote{\cite{kiros2013multimodal} further improved their results after the publication. We compare our results with their updated ones here.} are shown in Table \ref{tab:iaprtc_gen}.
% BACK-OFF GT2 and GT3 are n-grams methods with Katz backoff and Good-Turing discounting (\cite{chen2000survey}).
Ours-RNN-Base serves as a baseline method for our m-RNN model.
It has the same architecture as m-RNN except that it does not have the image representation input.

To conduct a fair comparison, we follow the same experimental settings of \cite{kiros2013multimodal} to calculate the BLEU scores and perplexity.
% , including the context length to calculate the BLEU scores and perplexity.
These two evaluation metrics are not necessarily correlated to each other for the following reasons.
As mentioned in Section \ref{sec:trainCost}, perplexity is calculated according to the conditional probability of the word in a sentence given all of its previous reference words.
Therefore, a strong language model that successfully captures the distributions of words in sentences can have a low perplexity without the image content.
But the content of the generated sentences might be uncorrelated to images.
From Table \ref{tab:iaprtc_gen}, we can see that although our baseline method of RNN generates a low perplexity, its BLEU score is low, indicating that it fails to generate sentences that are consistent with the content of images.

Table \ref{tab:iaprtc_gen} shows that our m-RNN model performs much better than our baseline RNN model and the state-of-the-art methods both in terms of the perplexity and BLEU score.
% It also outperforms the state-of-the-art methods in terms of perplexity, B-1, B-3, and a comparable result for B-2.
% \footnote{\cite{kiros2013multimodal} further improve their results after the publication. The perplexity of MLBL-F and LBL now are 9.90 and 9.29 respectively.}.

For the retrieval tasks, since there are no publicly available results of R@K and Med r in this dataset, we report R@K scores of our method in Table \ref{tab:iaprtc_ret} for future comparisons.
The result shows that 20.9\% top-ranked retrieved sentences and 13.2\% top-ranked retrieved images are groundtruth.
We also adopt additional evaluation metrics to compare our method with \cite{kiros2013multimodal}, see supplementary material Section \ref{supp:iapr_ret}.

\begin{table}[!tb]
	\centering
\begin{tabular}{l|ccccc}
\hline
      & $\mathcal{PPL}$  & B-1   & B-2   & B-3 & B-4 \\
\hline
% BACK-OFF GT2 & 54.5  & 0.323 & 0.145 & 0.059 \\
% BACK-OFF GT3 & 55.6  & 0.312 & 0.131 & 0.059 \\
LBL, \cite{mnih2007three} & 9.29  & 0.321 & 0.145 & 0.064 & - \\
MLBLB-AlexNet, \cite{kiros2013multimodal} & 9.86  & 0.393 & 0.211 & 0.112 & - \\
MLBLF-AlexNet, \cite{kiros2013multimodal} & 9.90  & 0.387 & 0.209 & 0.115 & - \\
\cite{gupta2012choosing} & -     & 0.15  & 0.06  & 0.01 & - \\
\cite{gupta2012image} & -     & 0.33  & 0.18  & 0.07 & - \\
\hdashline
Ours-RNN-Base & 7.77  & 0.307 & 0.177 & 0.096 & 0.043 \\
Ours-m-RNN-AlexNet & \textbf{6.92} & \textbf{0.482} & \textbf{0.357} & \textbf{0.269} & \textbf{0.208} \\
\hline
\end{tabular}%
\vspace{-0.2cm}
	\caption{Results of the sentence generation task on the IAPR TC-12 dataset. ``B'' is short for BLEU. }
	\label{tab:iaprtc_gen}
% \vspace{-0.2cm}
\end{table}

\begin{table}[!t]
	\centering
\begin{tabular}{l|cccc|cccc}
\hline
      & \multicolumn{4}{c|}{Sentence Retrival (Image to Text)} & \multicolumn{4}{c}{Image Retrival (Text to Image)} \\
      & R@1   & R@5   & R@10  & Med r & R@1   & R@5   & R@10  & Med r \\
\hline
Ours-m-RNN & 20.9  & 43.8  & 54.4  & 8     & 13.2  & 31.2  & 40.8  & 21 \\
\hline
\end{tabular}%
% \vspace{-0.2cm}
	\caption{R@K and median rank (Med r) for IAPR TC-12 dataset.}
	\label{tab:iaprtc_ret}
% \vspace{-0.5cm}
\end{table}

\subsection{Results on Flickr8K}

This dataset was widely used as a benchmark dataset for image and sentence retrieval.
The R@K and Med r of different methods are shown in Table \ref{tab:flickr8K_ret}.
We compare our model with several state-of-the-art methods: SDT-RNN (\cite{socher2014grounded}), DeViSE (\cite{frome2013devise}), DeepFE (\cite{karpathy2014fragment}) with various image representations.
Our model outperforms these methods by a large margin when using the same image representation (e.g. AlexNet).
We also list the performance of methods using more sophisticated features in Table \ref{tab:flickr8K_ret}.
``-avg-RCNN'' denotes methods with features of the average CNN activation of all objects above a detection confidence threshold.
DeepFE-RCNN \cite{karpathy2014fragment} uses a fragment mapping strategy to better exploit the object detection results.
The results show that using these features improves the performance.
Even without the help from the object detection methods, however, our method performs better than these methods in almost all the evaluation metrics.
We will develop our framework using better image features based on object detection in the future work.

\begin{table}[!t]
	\centering
\begin{tabular}{l|cccc|cccc}
\hline
      & \multicolumn{4}{c|}{Sentence Retrival (Image to Text)} & \multicolumn{4}{c}{Image Retrival (Text to Image)} \\
      & R@1   & R@5   & R@10  & Med r & R@1   & R@5   & R@10  & Med r \\
\hline
Random & 0.1   & 0.5   & 1.0   & 631   & 0.1   & 0.5   & 1.0   & 500 \\
SDT-RNN-AlexNet & 4.5   & 18.0  & 28.6  & 32    & 6.1   & 18.5  & 29.0  & 29 \\
Socher-avg-RCNN & 6.0   & 22.7  & 34.0  & 23    & 6.6   & 21.6  & 31.7  & 25 \\
DeViSE-avg-RCNN & 4.8   & 16.5  & 27.3  & 28    & 5.9   & 20.1  & 29.6  & 29 \\
DeepFE-AlexNet & 5.9   & 19.2  & 27.3  & 34    & 5.2   & 17.6  & 26.5  & 32 \\
DeepFE-RCNN & 12.6  & 32.9  & 44.0  & 14    & 9.7   & 29.6  & \textbf{42.5} & \textbf{15} \\
\hdashline
Ours-m-RNN-AlexNet & \textbf{14.5} & \textbf{37.2} & \textbf{48.5} & \textbf{11} & \textbf{11.5} & \textbf{31.0} & 42.4  & \textbf{15}\\
\hline
\end{tabular}%
	\caption{Results of R@K and median rank (Med r) for Flickr8K dataset.
	``-AlexNet'' denotes the image representation based on AlexNet extracted from the whole image frame.
	``-RCNN'' denotes the image representation extracted from possible objects detected by the RCNN algorithm.}
	\label{tab:flickr8K_ret}
\end{table}

The $\mathcal{PPL}$, B-1, B-2, B-3 and B-4 of the generated sentences using our m-RNN-AlexNet model in this dataset are 24.39, 0.565, 0.386, 0.256, and 0.170 respectively. 
% generated sentences in Table \ref{tab:flickr8_30K_gen} and compared our method with some of the recent proposed methods in this dataset.
%There is no publicly available algorithm that reported results on this dataset.
%So we compared our m-RNN model with the Ours-RNN-Base model.
%The m-RNN model performs much better than this baseline both in terms of the perplexity and BLEU scores.

\subsection{Results on Flickr30K and MS COCO}

\begin{table}[htb]
	\centering
\begin{tabular}{l|cccc|cccc}
\hline
      & \multicolumn{4}{c|}{Sentence Retrival (Image to Text)} & \multicolumn{4}{c}{Image Retrival (Text to Image)} \\
      & R@1   & R@5   & R@10  & Med r & R@1   & R@5   & R@10  & Med r \\
\hline
\multicolumn{9}{c}{Flickr30K}\\
\hline
Random & 0.1   & 0.6   & 1.1   & 631   & 0.1   & 0.5   & 1.0   & 500 \\
DeViSE-avg-RCNN & 4.8   & 16.5  & 27.3  & 28    & 5.9   & 20.1  & 29.6  & 29 \\
DeepFE-RCNN & 16.4  & 40.2 & 54.7 & 8 & 10.3  & 31.4 & 44.5 & 13 \\
RVR   & 12.1  & 27.8  & 47.8  & 11    & 12.7  & 33.1  & 44.9  & 12.5 \\
MNLM-AlexNet & 14.8  & 39.2  & 50.9  & 10    & 11.8  & 34.0  & 46.3  & 13 \\
MNLM-VggNet & 23.0  & 50.7  & 62.9  & 5     & 16.8  & 42.0  & 56.5  & 8 \\
NIC   & 17.0  & 56.0  & -     & 7     & 17.0  & \textbf{57.0} & -     & 7 \\
LRCN  & 14.0  & 34.9  & 47.0  & 11    & -     & -     & -     & - \\
DeepVS & 22.2  & 48.2  & 61.4  & 4.8   & 15.2  & 37.7  & 50.5  & 9.2 \\
\hdashline
Ours-m-RNN-AlexNet & 18.4  & 40.2  & 50.9  & 10    & 12.6  & 31.2  & 41.5  & 16 \\
Ours-m-RNN-VggNet & \textbf{35.4} & \textbf{63.8} & \textbf{73.7} & \textbf{3} & \textbf{22.8} & 50.7  & \textbf{63.1} & \textbf{5} \\
\hline
\multicolumn{9}{c}{MS COCO}\\
\hline
Random & 0.1   & 0.6   & 1.1   & 631   & 0.1   & 0.5   & 1.0   & 500 \\
DeepVS-RCNN & 29.4  & 62.0  & 75.9  & 2.5   & 20.9  & \textbf{52.8} & 69.2  & 4 \\
\hdashline
Ours-m-RNN-VggNet & \textbf{41.0} & \textbf{73.0} & \textbf{83.5} & \textbf{2} & \textbf{29.0} & 42.2  & \textbf{77.0} & \textbf{3} \\
\hline
\end{tabular}%
% \vspace{-0.3em}
	\caption{Results of R@K and median rank (Med r) for Flickr30K dataset and MS COCO dataset.}
	\label{tab:flickr30K_COCO_ret}
\end{table}

\begin{table}[htb]
	\centering
	\tabcolsep=0.15cm
\begin{tabular}{l|ccccc|ccccc}
\hline
      & \multicolumn{5}{c|}{Flickr30K}         & \multicolumn{5}{c}{MS COCO} \\
      & $\mathcal{PPL}$  & B-1   & B-2   & B-3   & B-4   & $\mathcal{PPL}$  & B-1   & B-2   & B-3   & B-4 \\
\hline
RVR   & -     & -     & -     & -     & 0.13  & -     & -     & -     & -     & 0.19 \\
DeepVS-AlexNet & -     & 0.47  & 0.21  & 0.09  & -     & -     & 0.53  & 0.28  & 0.15  & - \\
DeepVS-VggNet & 21.20 & 0.50  & 0.30  & 0.15  & -     & 19.64 & 0.57  & 0.37  & 0.19  & - \\
NIC   & -     & \textbf{0.66}  & -     & -     & -     & -     & \textbf{0.67}  & -     & -     & - \\
LRCN  & -     & 0.59  & 0.39  & 0.25  & 0.16  & -     & 0.63  & 0.44  & 0.31  & 0.21 \\
DMSM  & -     & -     & -     & -     & -     & -     & -     & -     & -     & 0.21 \\
\hdashline
Ours-m-RNN-AlexNet & 35.11 & 0.54  & 0.36  & 0.23  & 0.15  & -     & -     & -     & -     & - \\
Ours-m-RNN-VggNet & \textbf{20.72} & 0.60  & \textbf{0.41}  & \textbf{0.28}  & \textbf{0.19}  & \textbf{13.60} & \textbf{0.67}  & \textbf{0.49}  & \textbf{0.35}  & \textbf{0.25} \\
\hline
\end{tabular}%
\caption{Results of generated sentences on the Flickr 30K dataset and MS COCO dataset.}
\label{tab:flickr30K_COCO_gen}
\end{table}

\begin{table}[!tbh]
	\centering
\begin{tabular}{l|cccccc}
\hline
      & Our m-RNN & MNLM  & NIC   & LRCN  & RVR   & DeepVS\\
\hline
RNN Dim.  & 256 & 300   & 512   & 1000 ($\times 4$) & 100   & 300-600 \\
LSTM  & No & Yes   & Yes   & Yes   & No    & No  \\
\hline
\end{tabular}%
	\caption{Properties of the recurrent layers for the five very recent methods.
			LRCN has a stack of four 1000 dimensional LSTM layers.
			We achieves state-of-the-art performance using a relatively small dimensional recurrent layer.
			LSTM (\cite{hochreiter1997long}) can be treated as a sophisticated version of the RNN.}
	\label{tab:rnnSize}
\end{table}

We compare our method with several state-of-the-art methods in these two recently released dataset (Note that the last six methods appear very recently, we use the results reported in their papers): DeViSE (\cite{frome2013devise}), DeepFE (\cite{karpathy2014fragment}), MNLM (\cite{kiros2014unifying}), DMSM (\cite{fang2014captions}), NIC (\cite{vinyals2014show}), LRCN (\cite{donahue2014long}), RVR (\cite{chen2014learning}), and DeepVS (\cite{karpathy2014deep}).
The results of the retrieval tasks and the sentence generation task \footnote{We only select the word with maximum probability each time in the sentence generation process in Table \ref{tab:flickr30K_COCO_gen} while many comparing methods (e.g. DMSM, NIC, LRCN) uses a beam search scheme that keeps the best K candidates. The beam search scheme will lead to better performance in practice using the same model.} are shown in Table \ref{tab:flickr30K_COCO_ret} and Table \ref{tab:flickr30K_COCO_gen} respectively.
We also summarize some of the properties of the recurrent layers adopted in the five very recent methods in Table \ref{tab:rnnSize}.

Our method with VggNet image representation (\cite{simonyan2014very}) outperforms the state-of-the-art methods, including the very recently released methods, in almost all the evaluation metrics.
Note that the dimension of the recurrent layer of our model is relatively small compared to the competing methods.
It shows the advantage and efficiency of our method that directly inputs the visual information to the multimodal layer instead of storing it in the recurrent layer.
The m-RNN model with VggNet performs better than that with AlexNet, which indicates the importance of strong image representations in this task.
71\% of the generated sentences for MS COCO datasets are novel (i.e. different from training sentences).

\begin{table}[!tb]
	\centering
\begin{tabular}{l|ccccccc}
\hline
      & B1    & B2    & B3    & B4    & CIDEr & ROUGE\_L & METEOR \\
\hline
m-RNN-greedy-c5 & 0.668 & 0.488 & 0.342 & 0.239 & 0.729 & 0.489 & 0.221 \\
m-RNN-greedy-c40 & 0.845 & 0.730 & 0.598 & 0.473 & 0.740 & 0.616 & 0.291 \\
\hdashline
m-RNN-beam-c5 & 0.680 & 0.506 & 0.369 & 0.272 & 0.791 & 0.499 & 0.225 \\
m-RNN-beam-c40 & 0.865 & 0.760 & 0.641 & 0.529 & 0.789 & 0.640 & 0.304 \\
\hline
\end{tabular}%
	\caption{Results of the MS COCO test set evaluated by MS COCO evaluation server}
	\label{tab:test_MSCOCO}
\end{table}

We also validate our method on the test set of MS COCO by their evaluation server (\cite{capeval2015}).
The results are shown in Table \ref{tab:test_MSCOCO}.
We evaluate our model with greedy inference (select the word with the maximum probability each time) as well as with the beam search inference.
``-c5'' represents results using 5 reference sentences and ``-c40'' represents results using 40 reference sentences.

To further validate the importance of different components of the m-RNN model, we train several variants of the original m-RNN model and compare their performance.
In particular, we show that the two-layer word embedding system outperforms the single-layer version and the strategy of directly inputting the visual information to the multimodal layer substantially improves the performance (about 5\% for B-1).
Due to the limited space, we put the details of these experiments in Section \ref{sec:eff_comp} in the supplementary material after the main paper.

\section{Nearest Neighbor as Reference}
\label{sec:NNref}
Recently, \cite{devlin2015exploring} proposed a nearest neighbor approach that retrieves the captions of the $k$ nearest images in the training set, ranks these captions according to the consensus of the caption w.r.t. to the rest of the captions, and output the top ranked one.

Inspired by this method, we first adopt the m-RNN model with the transposed weight sharing strategy (\cite{mao2015learning}, denoted as m-RNN-shared) to generate $n$ hypotheses using a beam search scheme.
Specifically, we keep the $n$ best candidates in the sentence generation process until the model generates the end sign $w_\text{end}$.
These $n$ best candidates are approximately the $n$ most probable sentences generated by the model, and can be treated as the $n$ hypotheses.
In our experiments, we set $n=10$ since it gives us a diversified set of hypotheses without too much outliers on our validation set.
\footnote{If we directly output the top hypotheses generated by the model, then $n=5$ gives us the best performance. But if we want to rerank the hypotheses, then $n=10$ gives us a better result on the validation set.}

After generating the hypotheses of a target image, we retrieve its nearest neighbors in the image feature space on the training set (see details in Section \ref{sec:NNimg}).
Then we calculate the ``consensus'' scores (\cite{devlin2015language}) of the hypotheses w.r.t. to the groundtruth captions of the nearest neighbor images, and rerank the hypotheses according to these scores (see details in Section \ref{sec:CR}).

\subsection{Image features for the nearest neighbor image search}
\label{sec:NNimg}
\begin{figure}[tb!]
\begin{center}
\includegraphics[width=0.98\linewidth]{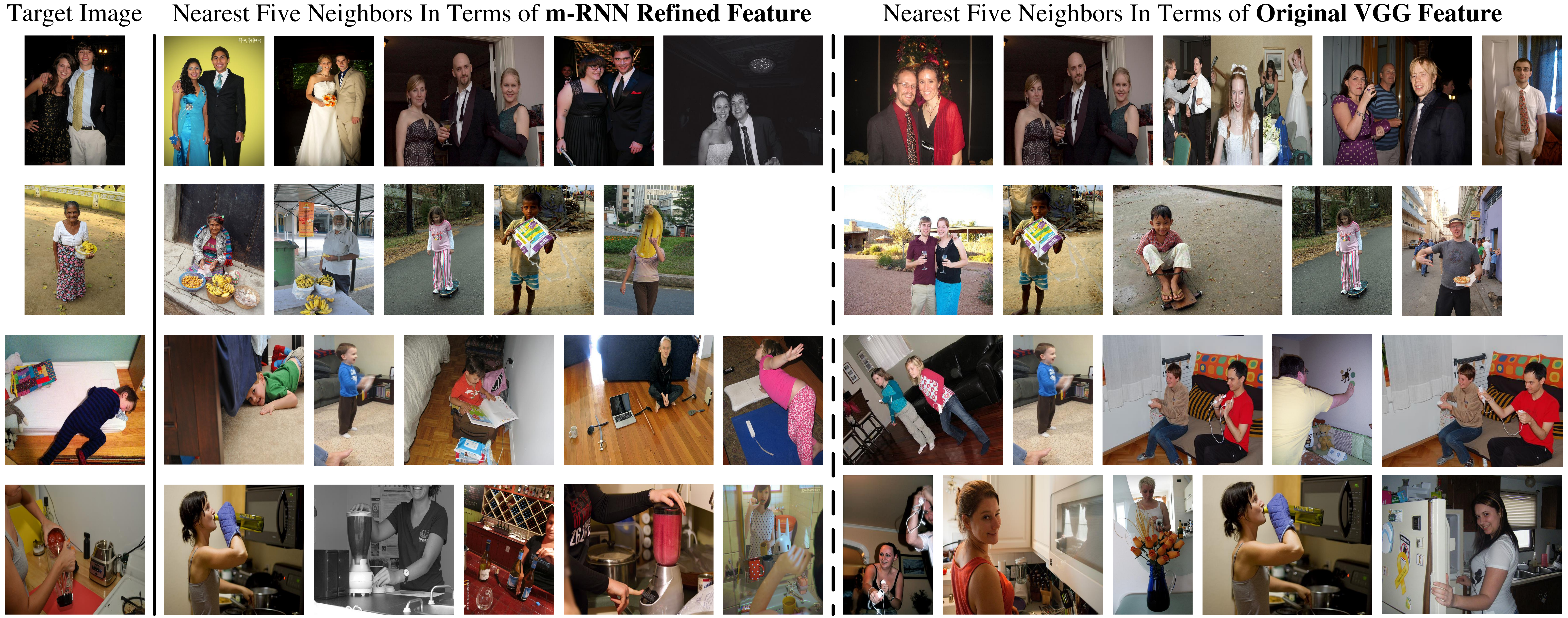}
\end{center}
   \caption{
            The sample images and their nearest neighbors retrieved by two types of features.
            Compared to the original VggNet features, the features refined by the m-RNN model are better for capturing richer and more accurate visual information. 
           }
\label{fig:NNimg}
\end{figure}

We try two types of image features for the nearest neighbor image search \footnote{We release both types of the features on MS COCO 2014 train, val and test sets. Please refer to the readme file at \url{https://github.com/mjhucla/mRNN-CR} to see how to download and use them.}.
The first one is the original image features extracted by the VggNet (\cite{simonyan2014very}).
We first resize the image so that its short side is 256 pixels.
Then we extract features on ten $224 \times 224$ windows (the four corners, the center and their mirrored versions) on the resized image.
Finally, we average pool the ten features to make it a 4,096 dimensional feature.

The second type is the feature refined by our m-RNN model.
It can be calculated as: $\mathbf{I}^r = g_2(\mathbf{V}_I \cdot \mathbf{I})$, where $\mathbf{V}_I$ is the weight matrix between the image representation and the multimodal layer (see Equation \ref{equ:multimodal}), and $g_2(.)$ is the scaled hyperbolic tangent function.

We show the sample images and their nearest neighbors in Figure \ref{fig:NNimg}.
We find that compared to the original VggNet features, the features refined by the m-RNN model capture richer and more accurate visual information. 
E.g., the target image in the second row contains an old woman with a bunch of bananas. 
The original VggNet features do not retrieve images with bananas in them.

\begin{table}[!b]
	\centering
	\tabcolsep=0.1cm
	\renewcommand{\arraystretch}{1.2}
\begin{tabular}{l|ccccccc}
\hline
\multicolumn{8}{c}{MS COCO val for consensus reranking} \\
\hline
      & B1    & B2    & B3    & B4    & CIDEr & ROUGE\_L & METEOR \\
\hline
m-RNN-shared & 0.686 & 0.511 & 0.375 & 0.280 & 0.842 & 0.500 & 0.228 \\
\hdashline
m-RNN-shared-NNref-BLEU & 0.718 & 0.550 & 0.409 & 0.305 & 0.909 & 0.519 & 0.235 \\
m-RNN-shared-NNref-CIDEr & 0.714 & 0.543 & 0.406 & 0.304 & 0.938 & 0.519 & 0.239 \\
\hdashline
m-RNN-shared-NNref-BLEU-Orcale & 0.792 & 0.663 & 0.543 & 0.443 & 1.235 & 0.602 & 0.287 \\
m-RNN-shared-NNref-CIDEr-Oracle & 0.784 & 0.648 & 0.529 & 0.430 & 1.272 & 0.593 & 0.287 \\
\hline
\multicolumn{8}{c}{MS COCO 2014 test server} \\
\hline
      & B1    & B2    & B3    & B4    & CIDEr & ROUGE\_L & METEOR \\
\hline
m-RNN-shared & 0.685 & 0.512 & 0.376 & 0.279 & 0.819 & 0.504 & 0.229 \\
\hdashline
m-RNN-shared-NNref-BLEU & 0.720 & 0.553 & 0.410 & 0.302 & 0.886 & 0.524 & 0.238 \\
m-RNN-shared-NNref-CIDEr & 0.716 & 0.545 & 0.404 & 0.299 & 0.917 & 0.521 & 0.242 \\
\hline
\end{tabular}%
	\caption{Results of m-RNN-shared model after applying consensus reranking using nearest neighbors as references (m-RNN-shared-NNref), compared with those of the original m-RNN model on our validation set and MS COCO test server.}
	\label{tab:res_cr_MSCOCO}
\end{table}

\subsection{Consensus Reranking}
\label{sec:CR}
Suppose we have get the $k$ nearest neighbor images in the training set as the reference.
We follow \cite{devlin2015language} to calculate the consensus score of a hypotheses.
The difference is that \cite{devlin2015language} treat the captions of the $k$ nearest neighbor images as the hypotheses while our hypotheses are generated by the m-RNN model.
More specifically, for each hypothesis, we calculate the mean similarity between this hypothesis and all the captions of the $k$ nearest neighbor images.
The consensus score of this hypothesis is the mean similarity score of the $m$ nearest captions.
The similarity between a hypothesis and one of its nearest neighbor reference captions is defined by a sentence-level BLEU score (\cite{papineni2002bleu}) or a sentence-level CIDEr (\cite{vedantam2014cider}).
We cross-validate the hyperparamters $k$ and $m$.
For the BLEU-based similarity, the optimal $k$ and $m$ are 60 and 175 respectively.
For the CIDEr-based similarity, the optimal $k$ and $m$ are 60 and 125 respectively.

\subsection{Experiments}

We show the results of our model on our validation set and the MS COCO testing server in Table \ref{tab:res_cr_MSCOCO}.
For BLEU-based consensus reranking, we get an improvement of 3.5 points on our validation set and 3.3 points on the MS COCO test 2014 set in terms of BLEU4 score.
For the CIDEr-based consensus reranking, we get an improvement of 9.4 points on our validation set and 9.8 points on the MS COCO test 2014 set in terms of CIDEr.

\subsection{Discussion}
\begin{figure}[b!]
\begin{center}
\includegraphics[width=0.98\linewidth]{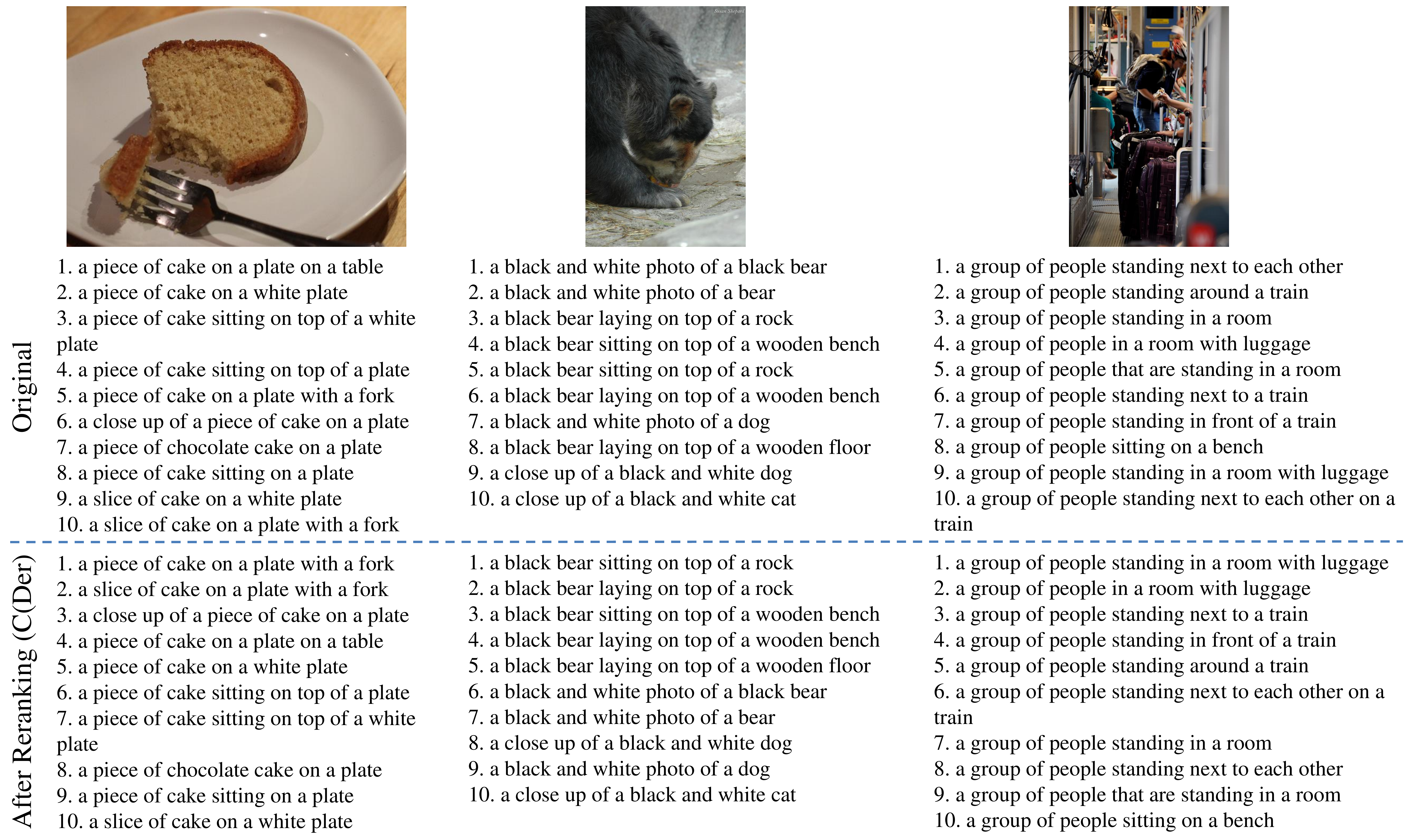}
\end{center}
   \caption{
            The original rank of the hypotheses and the rank after consensus reranking (CIDEr).
           }
\label{fig:vis_cr}
\end{figure}

We show the rank of the ten hypotheses before and after reranking in Figure \ref{fig:vis_cr}.
Although the hypotheses are similar to each other, there are some variances among them (E.g., some of them capture more details of the images. Some of them might be partially wrong).
The reranking process is able to improve the rank of good captions.

We also show the oracle performance of the ten hypotheses, which is the upper bound of the consensus reranking.
More specifically, for each image in our validation set, we rerank the hypotheses according to the scores (BLEU or CIDEr) w.r.t to the groundtruth captions.
The results of this oracle reranking are shown in Table \ref{tab:res_cr_MSCOCO} (see rows with ``-oracle'').
The oracle performance is surprisingly high, indicating that there is still room for improvement, both for the m-RNN model itself and the reranking strategy.

\section{Conclusion}

We propose a multimodal Recurrent Neural Network (m-RNN) framework that performs at the state-of-the-art in three tasks: sentence generation, sentence retrieval given query image and image retrieval given query sentence.
The model consists of a deep RNN, a deep CNN and these two sub-networks interact with each other in a multimodal layer.
Our m-RNN is powerful of connecting images and sentences and is flexible to incorporate more complex image representations and more sophisticated language models.

\subsubsection*{Acknowledgments}
We thank Andrew Ng, Kai Yu, Chang Huang, Duohao Qin, Haoyuan Gao, Jason Eisner for useful discussions and technical support. 
We also thank the comments and suggestions of the anonymous reviewers from ICLR 2015 and NIPS 2014 Deep Learning Workshop.
We acknowledge the Center for Minds, Brains and Machines (CBMM), partially funded by NSF STC award CCF-1231216, and ARO 62250-CS.

\bibliography{iclr2015}
\bibliographystyle{iclr2015}

\section{Supplementary Material}
\subsection{Effectiveness of the different components of the m-RNN model}
\label{sec:eff_comp}

\begin{figure}[bth]
\begin{center}
\includegraphics[width=0.85\linewidth]{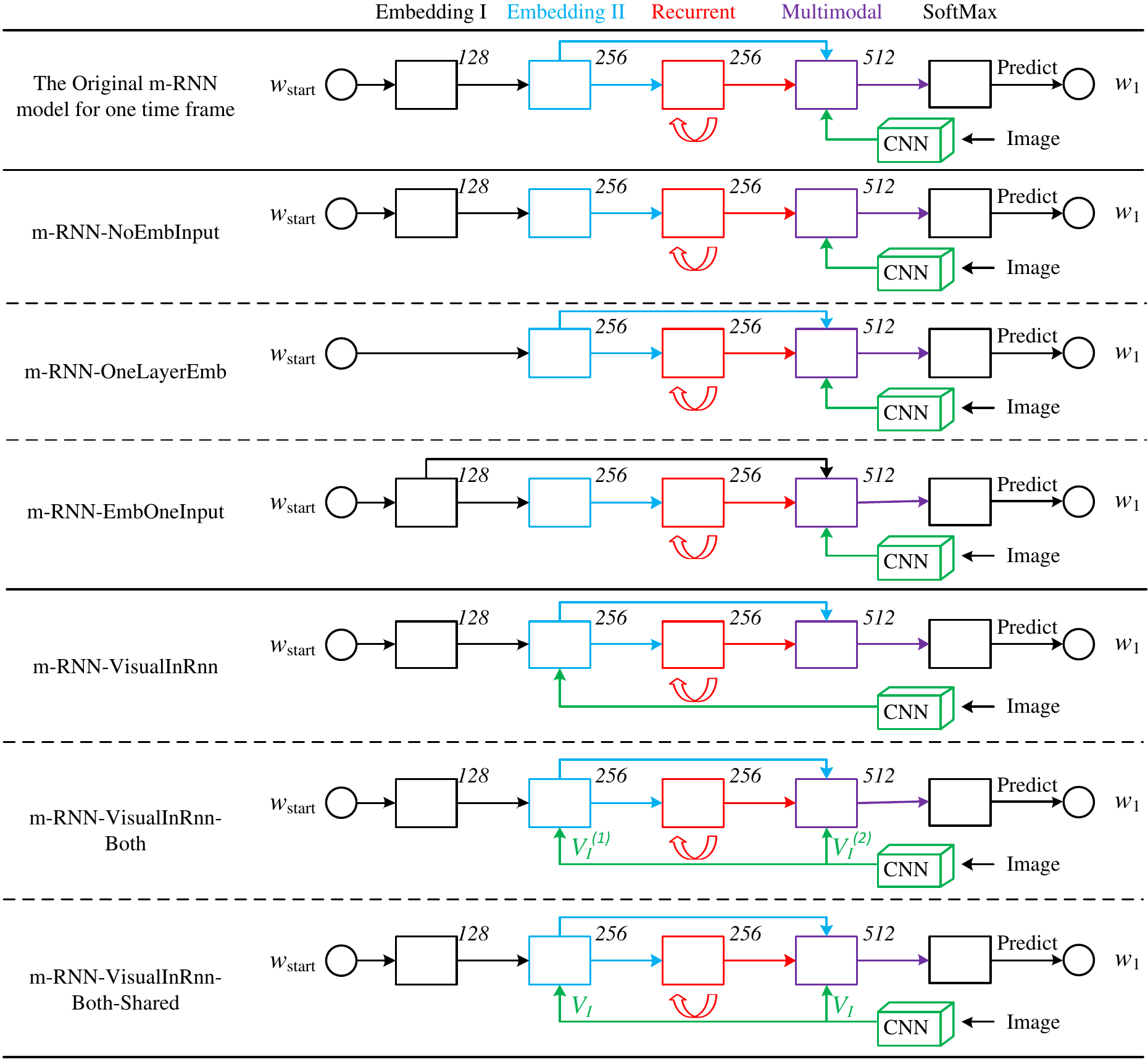}
\end{center}
   \caption{Illustration of the seven variants of the m-RNN models.
   }
\label{fig:eff_illu}
\end{figure}

\begin{table}[bth]
	\centering
\begin{tabular}{l|ccccc}
\hline
      & B-1   & B-2   & B-3   & B-4 \\
\hline
m-RNN & 0.600 & 0.412 & 0.278 & 0.187 \\
\hdashline
m-RNN-NoEmbInput & 0.592 & 0.408 & 0.277 & 0.188 \\
m-RNN-OneLayerEmb & 0.594 & 0.406 & 0.274 & 0.184 \\
m-RNN-EmbOneInput & 0.590 & 0.406 & 0.274 & 0.185 \\
\hdashline
m-RNN-visInRnn & 0.466 & 0.267 & 0.157 & 0.101 \\
m-RNN-visInRnn-both & 0.546 & 0.333 & 0.191 & 0.120 \\
m-RNN-visInRnn-both-shared & 0.478 & 0.279 & 0.171 & 0.110 \\
\hline
\end{tabular}%
\caption{Performance comparison of different versions of m-RNN models on the Flickr30K dataset.
All the models adopt VggNet as the image representation.
See Figure \ref{fig:eff_illu} for details of the models.}
\label{tab:eff_comp}
\end{table}

In this section, we compare different variants of our m-RNN model to show the effectiveness of the two-layer word embedding and the strategy to input the visual information to the multimodal layer.

\textbf{The word embedding system.}
Intuitively, the two word embedding layers capture high-level semantic meanings of words more efficiently than the single layer word embedding.
As an input to the multimodal layer, it offers useful information for predicting the next word distribution.

To validate its efficiency, we train three different m-RNN networks: m-RNN-NoEmbInput, m-RNN-OneLayerEmb, m-RNN-EmbOneInput.
They are illustrated in Figure \ref{fig:eff_illu}.
``m-RNN-NoEmbInput'' denotes the m-RNN model whose connection between the word embedding layer \rom{2} and the multimodal layer is cut off.
Thus the multimodal layer has only two inputs: the recurrent layer and the image representation.
``m-RNN-OneLayerEmb'' denotes the m-RNN model whose two word embedding layers are replaced by a single 256 dimensional word-embedding layer.
There are much more parameters of the word-embedding layers in the m-RNN-OneLayerEmb than those in the original m-RNN ($256 \cdot M$ v.s. $128 \cdot M + 128 \cdot 256$) if the dictionary size $M$ is large.
``m-RNN-EmbOneInput'' denotes the m-RNN model whose connection between the word embedding layer \rom{2} and the multimodal layer is replaced by the connection between the word embedding layer \rom{1} and the multimodal layer.
The performance comparisons are shown in Table \ref{tab:eff_comp}.

Table \ref{tab:eff_comp} shows that the original m-RNN model with the two word embedding layers and the connection between word embedding layer \rom{2} and multimodal layer performs the best.
It verifies the effectiveness of the two word embedding layers.

\textbf{How to connect the vision and the language part of the model.}
We train three variants of m-RNN models where the image representation is inputted into the recurrent layer: m-RNN-VisualInRNN, m-RNN-VisualInRNN-both, and m-RNN-VisualInRNN-Both-Shared.
For m-RNN-VisualInRNN, we only input the image representation to the word embedding layer \rom{2} while for the later two models, we input the image representation to both the multimodal layer and word embedding layer \rom{2}.
The weights of the two connections $V_I^{(1)}$, $V_I^{(2)}$ are shared for m-RNN-VisualInRNN-Both-Shared.
Please see details of these models in Figure \ref{fig:eff_illu}.
Table \ref{tab:eff_comp} shows that the original m-RNN model performs much better than these models, indicating that it is effective to directly input the visual information to the multimodal layer.

In practice, we find that it is harder to train these variants than to train the original m-RNN model and we have to keep the learning rate very small to avoid the exploding gradient problem.
Increasing the dimension of the recurrent layer or replacing RNN with LSTM (a sophisticated version of RNN \cite{hochreiter1997long}) might solve the problem.
We will explore this issue in future work.

\subsection{Additional retrieval performance comparisons on IAPR TC-12}
\label{supp:iapr_ret}

\begin{figure}[!b]
        \centering
        \begin{subfigure}[b]{0.42\textwidth}
                \includegraphics[width=\textwidth]{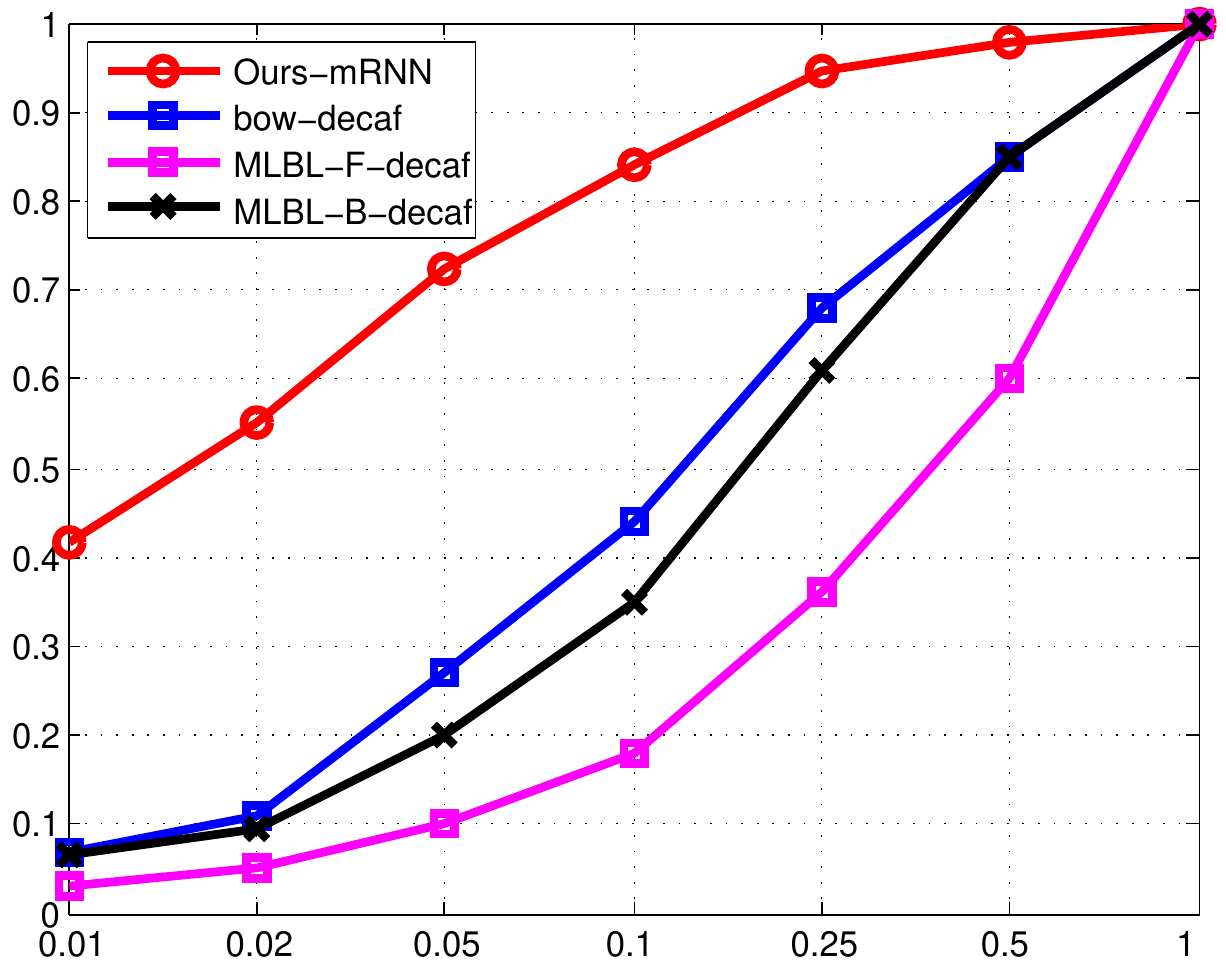}
                \caption{Image to Text Curve}
        \end{subfigure}%
        ~ %add desired spacing between images, e. g. ~, \quad, \qquad, \hfill etc.
        \begin{subfigure}[b]{0.42\textwidth}
                \includegraphics[width=\textwidth]{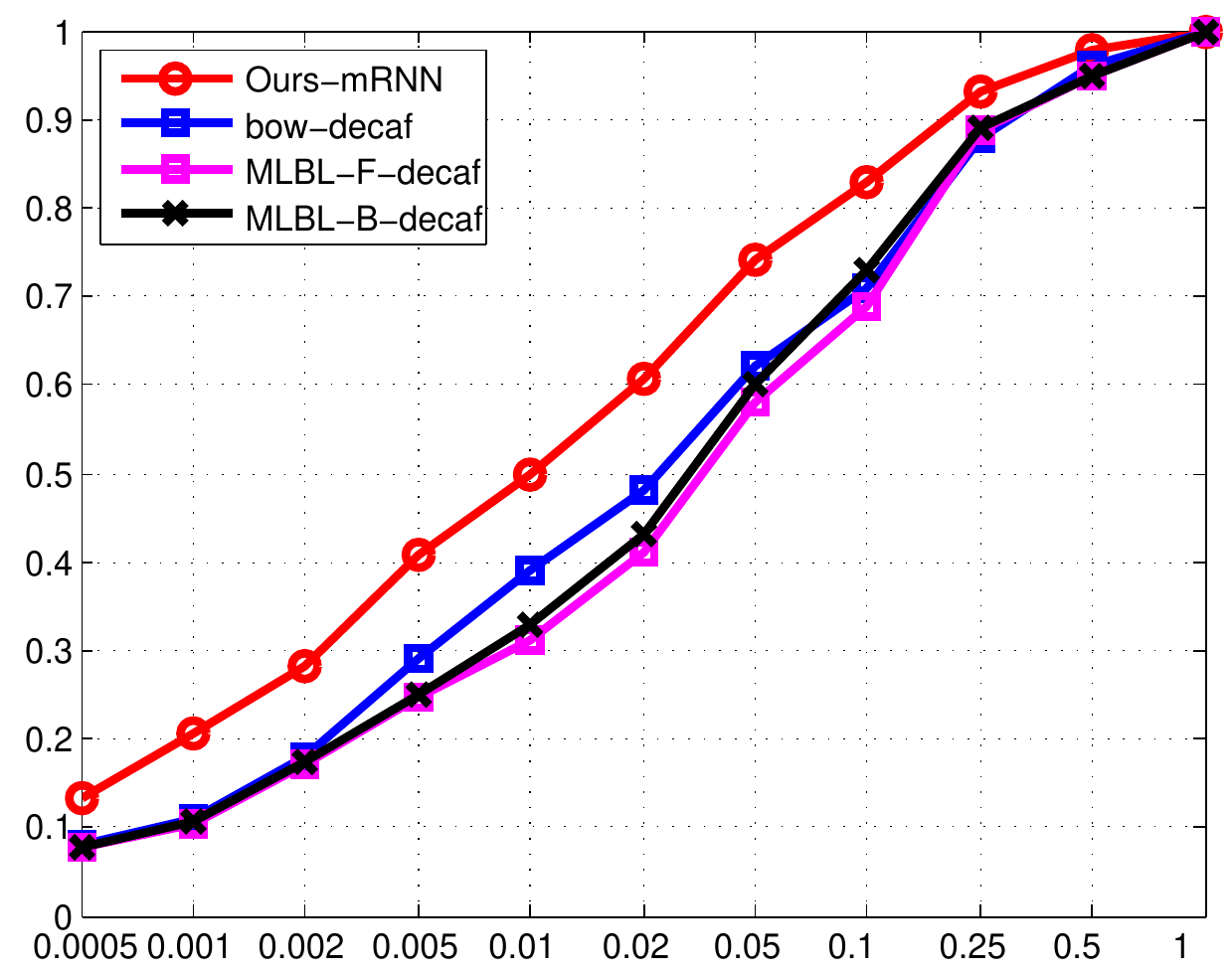}
                \caption{Text to Image Curve}
        \end{subfigure}
        \caption{Retrieval recall curve for (a). Sentence retrieval task (b). Image retrieval task on IAPR TC-12 dataset. The behavior on the far left (i.e. top few retrievals) is most important.}
        \label{fig:iaprtc_ret_curve}
\end{figure}

For the retrieval results in this dataset, in addition to the R@K and Med r, we also adopt exactly the same evaluation metrics as \cite{kiros2013multimodal} and plot the mean number of matches of the retrieved groundtruth sentences or images with respect to the percentage of the retrieved sentences or images for the testing set.
For the sentence retrieval task, \cite{kiros2013multimodal} uses a shortlist of 100 images which are the nearest neighbors of the query image in the feature space.
This shortlist strategy makes the task harder because similar images might have similar descriptions and it is often harder to find  subtle differences among the sentences and pick the most suitable one.
% Although there are no published R@K scores and Med r score for this dataset available for the best of our knowledge, we also report these scores of our method for future comparison.

The recall accuracy curves with respect to the percentage of retrieved images (sentence retrieval task) or sentences (sentence retrieval task) are shown in Figure \ref{fig:iaprtc_ret_curve}.
The first method, bow−decaf, is a strong image based bag-of-words baseline (\cite{kiros2013multimodal}).
The second and the third models (\cite{kiros2013multimodal}) are all multimodal deep models.
Our m-RNN model significantly outperforms these three methods in this task.

\subsection{The calculation of BLEU score}
\label{supp:bleu}
The BLEU score was proposed by \cite{papineni2002bleu} and was originally used as a evaluation metric for machine translation.
To calculate BLEU-N (i.e. B-N in the paper where $N$=1,2,3,4) score, we first compute the \emph{modified} n-gram precision (\cite{papineni2002bleu}), $p_n$.
Then we compute the geometric mean of $p_n$ up to length $N$ and multiply it by a brevity penalty BP:
\begin{equation}
\text{BP} = \min (1, e^{1-\frac{r}{c}})
\end{equation}
\begin{equation}
\text{B-N} = \text{BP} \cdot e^{\frac{1}{N}\sum_{n=1}^{N} \log p_n}
\end{equation}
where $r$ is the length of the reference sentence and $c$ is the length of the generated sentence.
We use the same strategy as \cite{fang2014captions} where $p_n$, $r$, and $c$ are computed over the whole testing corpus.
When there are multiple reference sentences, the length of the reference that is closest (longer or shorter) to the length of the candidate is used to compute $r$.

\end{document}